%% file: main.tex
\documentclass[runningheads]{llncs}
\usepackage{graphicx}
\usepackage{multirow}
\usepackage{booktabs}
\usepackage{makecell}
\usepackage{csquotes}
\usepackage{tikz}
\usepackage{comment}
\usepackage{enumerate}
\usepackage{amsmath,amssymb} % define this before the line numbering.
\usepackage{color}
\usepackage{soul}
\PassOptionsToPackage{hyphens}{url}\usepackage[breaklinks=true,bookmarks=falsel,colorlinks=true,linkcolor=black,anchorcolor=black,citecolor=black,filecolor=black,menucolor=black,runcolor=black,urlcolor=black]{hyperref}
\usepackage[accsupp]{axessibility}  % Improves PDF readability for those with disabilities.

\begin{document}
\input{definitions}

% \linenumbers
\pagestyle{headings}
\mainmatter
\def\ECCVSubNumber{2104}  % Insert your submission number here

\newcommand{\etal}{\textit{et al}.}

\title{Target-absent Human Attention}

% \title{Predicting Human Attention When There is No Target}

\begin{comment}
\titlerunning{ECCV-22 submission ID \ECCVSubNumber}
\authorrunning{ECCV-22 submission ID \ECCVSubNumber} 
\author{Anonymous ECCV submission}
\institute{Paper ID \ECCVSubNumber}
\end{comment}
%******************

% \begin{comment}
\titlerunning{Target-absent Human Attention}
\author{Zhibo Yang\and
Sounak Mondal\and
Seoyoung Ahn\and\\
Gregory Zelinsky\and
Minh Hoai\and
Dimitris Samaras
}

\authorrunning{Z. Yang et al.}

\institute{Stony Brook University, Stony Brook, NY 11794, USA}
% \end{comment}
%******************
\maketitle

\begin{abstract}
The prediction of human gaze behavior is important for building human-computer interaction systems that can anticipate the user's attention. Computer vision models have been developed to predict the fixations made by people as they search for target objects. But what about when the target is not in the image? Equally important is to know how people search when they cannot find a target, and when they would stop searching. In this paper, we propose a data-driven computational model that addresses the search-termination problem and predicts the scanpath of search fixations made by people searching for targets that do not appear in images. We model visual search as an imitation learning problem and represent the internal knowledge that the viewer acquires through fixations using a novel state representation that we call {\it Foveated Feature Maps (FFMs)}. FFMs integrate a simulated foveated retina into a pretrained ConvNet that produces an in-network feature pyramid, all with minimal computational overhead. Our method integrates FFMs as the state representation in inverse reinforcement learning. Experimentally, we improve the state of the art in predicting human target-absent search behavior on the COCO-Search18 dataset. 
Code is available at: \url{https://github.com/cvlab-stonybrook/Target-absent-Human-Attention}.
% \footnote{Code: \url{https://github.com/cvlab-stonybrook/Target-absent-Human-Attention}.}

\end{abstract}

% \section{Framing}
% Key questions this paper aims to answer:
% \begin{itemize}
%     \item How to predict human attention in the absence of a target?
%     \begin{itemize}
%         \item What's special about target-absent prediction?
%         \item Why existing approaches do not work well?
%         \item How do we address the problem in existing methods?
%     \end{itemize}
%     \item When does the search stop?
%     \begin{itemize}
%         \item What matters in termination criteria (reward, value, or time)?
%     \end{itemize}
%     \item Do subjects differ much in the target-absent searching behavior and their stopping criterion?
%     \item Are humans greedy searcher or long-term or short-term planner?
% \end{itemize}
% What's on the table?
% \begin{itemize}
%     \item Foveated features
%     \item Target-absent scanpath prediction
%     \item termination prediction
%     \item context-target analysis
%     \item IQ-Learn
%     \item semantic sequence score
% \end{itemize}
% Todos
% \begin{itemize}
%     \item \st{complete method section}
%     \item \st{Set up framework in experiment section}
%     \item \st{add semantic sequence score, step-wise saliency metrics}
%     \item \st{add main results table}
%     \item \st{add CVPR21 and other heuristics comparison}
%     \item \st{complete the result tables}
%     \item analyze time vs return in termination criterion 
%     \item re-draw Fig 1 and Fig 2.
% \end{itemize}

\section{Introduction}
The attention mechanism used by humans to prioritize and select visual information~\cite{posner1990attention,posner1994attention,petersen2012attention} has  attracted the interest of computer vision researchers seeking to reproduce this selection efficiency in machines \cite{yang2020predicting,chen2021coco,zelinsky2019benchmarking,chen2021predicting,rashidi2020optimal}. The most often-used paradigm to study this efficiency is a visual search task, where efficiency is measured with respect to how many attention shifts (gaze fixations) are needed to detect a target in an image. But what about when the target is not there? Understanding gaze behavior during target-absent search (including search termination) would serve applications in human-computer interaction while addressing basic questions in attention research. No predictive model of human search fixations would be complete without addressing the unique problems arising from target-absent search.

The neuroanatomy of the primate foveated retina is such that visual acuity decreases with increasing distance from the high-resolution central fovea. 
When searching for a target, this foveated retina drives people to move their eyes selectively to image locations most likely to be the target, thereby providing the highest-resolution visual input to the target-recognition task, with each fixation movement guided by low-resolution input from peripheral vision. 
% \young{When searching for a target, this foveated retina drives people to move their eyes. Each movement of fixation guided by low-resolution input from peripheral vision so that the fovea processes the interesting and important parts of the image (where the target is most likely to appear) at the highest-possible resolution.}
%This drives people to move their eyes to selectively look at  high-resolution pixels (guided by the low-resolution peripheral vision) in searching for the target. 
Recognizing the fact that the human visual input is filtered through a foveated retina is crucial to understanding and predicting human gaze behavior, and this is especially true for target-absent search where there is no clear target signal and gaze is driven instead by contextual relationships to other objects and the spatial cues that might provide about the target's location. %Therefore, it is crucial to simulate the foveated retina in a computational model in order to understand and predict human gaze behavior. This is especially important for target-absent search because no clear target signal is present and gaze is largely driven by context objects which provide spatial cues for locating the target.

To simulate a foveated retina for predicting human search fixations, Zelinsky \etal~\cite{zelinsky2019benchmarking} directly applied a pretrained ResNet \cite{he2016deep} to foveated images \cite{perry2002gaze} to extract feature maps for the state representation. Yang \etal~\cite{yang2020predicting} proposed DCBs that approximate a high-resolution fovea and a low-resolution periphery by using the segmentation maps of a full-resolution image and its blurred version, respectively, predicted by a pretrained Panoptic-FPN \cite{kirillov2019panopticfpn}.
% combining the panoptic segmentation maps of a pair of full-resolution and blurred images at the fixated locations using a pretrained Panoptic-FPN \cite{kirillov2019panopticfpn}.
Like other models for predicting human attention \cite{linardos2021deepgaze,kummerer2014deep,kummerer2015information,chen2021predicting,zelinsky2021predicting}, these approaches rely on pretrained networks to extract image features and train much smaller networks for the downstream tasks using transfer learning, usually due to the lack of human fixation data for training. Also noteworthy is that these approaches apply networks pretrained on full-resolution images (e.g., ResNets \cite{he2016deep} trained on ImageNet \cite{russakovsky2015imagenet}) on blurred images, 
expecting the pretrained networks to approximate how humans perceive blurred images. 
% However, studies \cite{hendrycks2018benchmarking,geirhos2018imagenettrained} have shown the opposite, that pretrained neural networks can be easily fooled by noisy images that include blur. 
{However, Convolutional Neural Networks (ConvNets) are highly vulnerable to image perturbation \cite{hendrycks2018benchmarking,geirhos2018imagenettrained} and the visual features extracted from the model on blurred images are hardly meaningful in the context of object recognition (contrary to human vision that actively seeks guidance from low-resolution peripheral vision for target recognition)}.

To better represent the degraded information that humans have available from their peripheral vision and can therefore use to guide their search, we exploit %To better capture the guidance of visual signals in the periphery for target-absent search and avoid the cumbersome creation of datasets and training detectors for individual targets in \cite{rashidi2020optimal}, we leverage
the fact that modern ConvNets have an inherent hierarchical architecture such that deeper layers have progressively larger receptive fields, corresponding to the greater blurring that occurs with increasing visual eccentricity. %(corresponding to larger eccentricity in the peripheral vision) and propose to combine
We propose combining the feature maps at different layers in a manner that is contingent upon the human fixation locations, approximating the information available from a foveated retina\footnote{Note that it is not our aim to perfectly approximate the information extracted by a human foveated retina.}. %in a foveated manner contingent on human 
We name this method {\it Foveated Feature Maps (FFMs)}. FFMs are computed on full-resolution images, so they can be readily applied to a wide range of pretrained ConvNets. Moreover, FFMs are a lightweight modification of modern ConvNets that capable of representing the subtle transition from fovea to periphery and are thus better suited for predicting human gaze movement. We find that our FFMs, when combined with inverse reinforcement learner (IQ-Learn \cite{garg2021iq}), significantly outperforms DCBs \cite{yang2020predicting} and other baselines (see \Sref{sec:main_result}) in predicting both target-absent and target-present fixations.
% Note that FFMs also avoid the need to create specialized datasets and train detectors for individual search targets, as required in \cite{rashidi2020optimal}.
% The capability of capturing object features in peripheral vision is especially important in target-absent search as no clear target signal is present and gaze is largely driven by context objects which provide spatial cues for locating the target. 

In short, our paper makes the following contributions: (1) 
% we introduce the first computational model for human attention prediction in target-absent search;
we introduce a data-driven computational model applicable to both the target-present and target-absent search prediction problems;
(2) we propose a new state representation that dynamically integrates knowledge collected via a foveated retina, similar to humans; (3) we predict target-absent search fixations at the ceiling of human performance, and achieve superior performance in predicting target-present scanpaths compared to previous methods; and (4) We propose a novel evaluation metric called {semantic sequence score} that measures the object-level consistency between human scanpaths. Compared to the traditional sequence score \cite{borji2013analysis}, it better captures the contextual cues that people use to guide their target-absent search behavior.
%Our contribution is summarized as follows
%\begin{enumerate}[1{)}]
%    \item We are the first to develop a computational model to predict target-absent human attention in visual search.
%    \item In order to predict human attention in target-absent tasks when no target signal is available, We propose a new state presentation called foveated feature maps that dynamically integrating knowledge collected via a foveated retina like humans do.
%    \item When combine with an inverse reinforcement learning algorithm, IQ-Learn \cite{garg2021iq}, our approach achieves superior performance in predicting target-absent human attention compared with previous methods.
%    \item We also propose a novel evaluation metric called semantic sequence score that measures the object-level consistency between human scanpaths. Compared with traditional sequence score \cite{yang2020predicting}, It better reflects the context guidance human leverages in target-absent search.
%    \item Despite designed for target-absent attention prediction, FFM is also readily applicable to target-present scenario. We show that our FFM also achieve the state-of-the-art performance in predicting target-present human attention.
%\end{enumerate}

\section{Related Work}

Visual search is one of the fundamental human goal-directed gaze behaviors that actively scan the visual environment to find any exemplar of a target-object category~\cite{wolfe1998visual,zelinsky2008theory,eckstein2011visual}. There is an emerging interest in modeling and predicting human gaze during visual search~\cite{yang2020predicting,zelinsky2019benchmarking,chen2021predicting,rashidi2020optimal,Chen_2022_CVPR}. Yang \etal~\cite{yang2020predicting} first used inverse reinforcement learning to model target-present search fixations spanning 18 target categories. Most recently, \cite{chen2021predicting} directly applied reinforcement learning to predict scanpaths in various visual tasks including target-present search. However, their generalizability has never been interrogated for the prediction of target-absent search scanpaths, where no strong target signal is available in the images. Early work showed that target-absent search is not random behavior~\cite{chun1996just,alexander2011visual} but greatly influenced by target-relevant visual features to such an extent that the target category being searched for can be decoded from one's scanpaths~\cite{zelinsky2013eye}. However, that study used only two target categories and the task was to search through only four non-targets. In this work, we study target-absent gaze behavior from a data-driven perspective.

% \myheading{Foveated retina modeling for scanpath prediction.}
% Several recent studies have engaged the question of how to model the human foveated retina for fixation prediction. 
Several recent studies have attempted to model a foveated representation of the input image for predicting human gaze behavior \cite{yang2020predicting,zelinsky2019benchmarking,rashidi2020optimal} or solving other visual tasks (e.g., object detection; \cite{akbas2017object,jaramillo2019foveated}).
Yang \etal~\cite{yang2020predicting} approximated a foveated retina by having a high-resolution center (full-resolution image) surrounded by a degraded visual periphery (a slightly blurred version of the image) at each fixation. A pretrained Panoptic-FPN \cite{kirillov2019panopticfpn} was applied on the full-resolution and blurred images separately to obtain the panoptic segmentation maps that were finally combined into the final state representation. Instead of approximating the foveated retina as a central-peripheral pairing of high- and low-resolution images, Zelinsky \etal~\cite{zelinsky2019benchmarking} used a pretrained ResNet-50 \cite{he2016deep} directly to extract feature maps from foveated images \cite{perry2002gaze} for the state representation. Notably, both methods apply pretrained networks on blurred images whereas our FFMs are extracted from the full-resolution images, for which the pretrained networks are more robust. 
% Rashidi \etal~\cite{rashidi2020optimal} attempted to better estimate the peripheral degradation of a foveated retina by modeling the foveated detectability of an object target \cite{najemnik2005optimal} measured as the discriminability between the feature distribution of target-present and target-absent images. 
Rashidi \etal~\cite{rashidi2020optimal} proposed a method to directly estimate the foveated detectability of a target object \cite{najemnik2005optimal} from eye tracking data.
% However, not only does this method require training multiple detectors for each target, it also requires each detector to be trained at different eccentricities, all using manually created datasets showing the targets at multiple scales against different textured backgrounds. This approach therefore cannot be easily extended to a larger number of target categories, as the 18 used in COCO-Search18 \cite{chen2021coco}. 
However, this approach cannot be easily extended to a larger number of target categories because it requires training multiple detectors for each target and manually creating specialized datasets by showing each target at multiple scales against different textured backgrounds.
In contrast, our model is able to jointly learn the foveation process at feature level and the networks that predict human scanpaths through back-propagation from human gaze behavior. 

\section{Approach}
Following the model of Yang \etal~\cite{yang2020predicting} for target-present data, we also propose to model visual search behavior for target absent data using  Inverse Reinforcement Learning (IRL). Specifically, we assume a human viewer is a reinforcement learning  agent trying to localize the target object on a given {\it target-absent} image (the human viewer does not know if the image contains the target or not).  The viewer acquires knowledge through a sequence of gaze fixations and allocates their next gaze point based on this knowledge  to search for the target (\Sref{sec:state}). The search is terminated when the viewer confirms there is no target in the given image (\Sref{sec:stop}). In this framework, we assume access to ground-truth human scanpaths (expert demonstrations), and the goal is to learn a policy that mimics or predicts human gaze behavior given an image and the target (\Sref{sec:policy}).

\subsection{Foveated Feature Maps (FFMs)}\label{sec:state}
\begin{figure}[t]
  \centering
  \includegraphics[width=1.\linewidth]{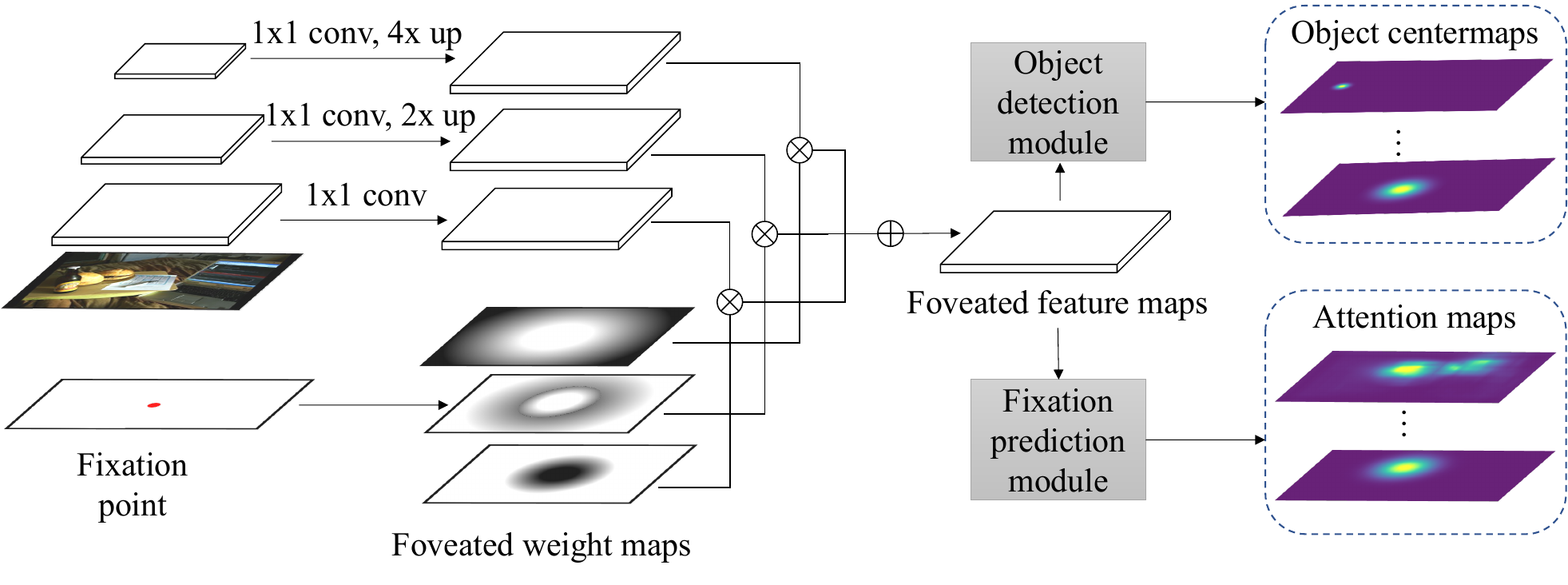}
  \caption{{\bf Overview of the foveated feature maps (FFMs)}. FFMs are a set of multi-resolution feature maps constructed by combining the in-network feature pyramid produced by a pretrained ConvNet using the foveated weight maps computed based on previous fixations. 
%   High-resolution feature maps are more weighted for locations closer to the fixation points. 
  The object detection module and the fixation prediction module map FFMs to a set of object center heatmaps (80 object categories in COCO \cite{lin2014microsoft}) and a set of attention maps for the 18 targets in COCO-Search18 \cite{chen2021coco}, respectively.}
  \label{fig:ffm}
\end{figure}

To capture the information a person acquires from an image through a sequence of fixations, we propose a novel state representation, called {Foveated Feature Map (FFMs)}. \Fref{fig:ffm} shows an overview of how our FFMs are constructed. FFMs take advantage of pretrained ConvNets, which produce a pyramid of feature maps with progressively larger receptive fields. By treating deeper feature maps as information obtained at larger eccentricity (lower-resolution) in the peripheral vision, we construct FFMs as a set of multi-resolution feature maps, which is a weighted combination of different levels of feature maps using foveated weight maps generated based on previous fixations.
Similar to image foveation \cite{zelinsky2019benchmarking}, in FFMs, deeper feature maps with a lower resolution (corresponding to a larger eccentricity) are more weighted at locations with increasing distance from the fixation points.
% Features with a larger receptive field (corresponding to a larger eccentricity) are more weighted with increasing distance from the fixated locations. 
Below we discuss FFMs in greater detail.
%\mhoai{I don't understand the last sentence.}

\myheading{Relative resolution map. } The human vision system is known to be foveated, meaning the visual information in the view is not processed at a uniform resolution. Rather, high spatial details are only obtained around the fixation point (i.e., the fovea) and the resolution outside of the fovea drops off as the distance between the peripheral pixels and the fovea increases. To simulate this, Perry and Geisler~\cite{perry2002gaze} proposed an image foveation method, which has been used in both free viewing \cite{jiang2015salicon} and visual search \cite{zelinsky2019benchmarking} tasks. Here, we extend image foveation to produce multi-resolutional feature maps to represent the foveated view of an image at the level of image features. Specifically, given a fixation $f=(x_f, y_f)$, we first define a relative resolution map contingent on $f$ as 
\begin{align}R(x,y|f)=\frac{\alpha}{\alpha + \frac{\sqrt{(x-x_f)^2 + (y-y_f)^2}}{p}}\cdot
\end{align}
Here, $p$ is the number of pixels in one degree of visual angle, which depends on the distance between the viewer and the display.  $\alpha$ is a learnable parameter that controls the decreasing speed of resolution as the pixel $(x,y)$ moves away from the fixation point.
%$\theta(x,y|f)$ denotes the visual eccentricity (i.e., visual angle) and is defined by the Euclidean distance between the pixel $(x,y)$ and the fixation point $f$: 
%\begin{align} 
%\theta(x,y|f) = \sqrt{(x-x_f)^2 + (y-y_f)^2}\Big/p,
%\end{align}
% where $p$ is the number of pixels in one degree of visual angle, depending on the distance between the viewer and the image display. 
% 

 For multiple fixations $\{f_1, \cdots,f_n\}$, we compute the combined resolution map by taking the maximum at every location:  $R(x,y|\{f_1, \cdots,f_n\}) = \max_{i} R(x,y|f_i)$. In contrast to \cite{perry2002gaze}, which creates a Gaussian pyramid of the given image $I$ to produce the multi-resolutional version of $I$, we take inspiration from the Feature Pyramid Network \cite{lin2017feature} and use the in-network feature pyramid produced by existing pretrained ConvNets and blend the feature maps at each level of the feature pyramid to construct multi-resolutional feature maps (i.e., FFMs) based on the relative resolution map $R(x,y|f)$. For brevity, we will write $R(x,y|f)$ as $R(x,y)$ in the following text.

\myheading{Foveated feature maps (FFMs). }
We use a ResNet-50 \cite{he2016deep} as the backbone (the method can be easily extended to other ConvNet backbones such as VGG nets \cite{simonyan2014very}), and let the feature pyramid from the ResNet be $\{C_1 \cdots, C_5\}$, which represents the feature activation outputs from the last residual block at each stage of ResNet-50, namely the outputs of conv1, conv2, conv3, conv4, and conv5. Similar to the Gaussian pyramid of an image, a lower level of the feature pyramid contains more spatial details, while  a higher-level feature map is stronger in semantics. To reduce the semantic discrepancy among different levels, we apply an $1{\times}1$ convolutional layer on every $C_i$ to project them to the same embedding space. Then, we upsample $\{C_i\}_{i=1}^5$ to the same spatial dimensions of $C_1$, yielding 3D tensors of the same size,  denoted as $\{P_1, \cdots, P_5\}$. We then compute a spatial weight map $W_i$ for each $P_i$ and produce a set of multi-resolution feature maps $M$ as the weighted combination of $W_i$ and $P_i$: $M=\sum_i W_i\odot P_i$, where $\odot$ denotes the element-wise multiplication at the spatial axes. We call these multi-resolution feature maps FFMs. Below we describe how to compute $W_i$ based on the relative resolution map $R(x,y)$.

Each level of the feature pyramid $P_i$ represents a certain eccentricity, corresponding to a fixed spatial resolution, which we denote as $R_i^*$. It is defined as the relative resolution where a transfer function $T_i(\cdot)$  is at its half maximum, i.e., $T_i(R_i^*)=0.5$ \cite{perry2002gaze}.
The transfer function $T_i(\cdot)$ is the function that maps relative resolution $r$ to relative amplitude, and it is defined as: 
\begin{align}T_i(r)=\exp(-({2^{i-3}r}/{\sigma})^2/2).
\end{align}
It can be shown that $R_1^*>R_2^*>R_3^*>R_4^*>R_5^*$, forming four resolution bins whose boundaries are defined by $R_i^*$ and $R_{i-1}^*$ ($i\in \{2,3,4,5\}$). To compute the weights at location $(x,y)$, we first determine which bin pixel $(x,y)$ falls in, according to its relative resolution $R(x,y)$ (see the supplementary material for more details). Assume pixel $(x,y)$ falls in between layer $j$ and $j-1$, i.e., $R_{j-1}^* \geq R(x,y)>R_{j}^*$. Then, we set the weights at layer $j$ and $j-1$ to be the ratio of the distance between pixel $(x,y)$ and the corresponding layer to the distance between the layer $j$ and $j-1$ at $(x,y)$ in relative amplitude space:
\begin{align} W_i(x,y) = 
\begin{cases} 
  \frac{0.5-T_j(R(x,y))}{T_{j-1}(R(x,y))-T_j(R(x,y))} & \textrm{if } i=j-1, \\
  1-\frac{0.5-T_j(R(x,y))}{T_{j-1}(R(x,y))-T_j(R(x,y))} & \textrm{if } i=j, \\
  0 & \text{otherwise.}
\end{cases} %\quad \forall i\in\{1,2,3,4\}
\end{align}
Apparently, $\sum_i W_i(x,y)=1$ and at location $(x,y)$ only features from layer $j$ and layer $j-1$ are integrated into the final FFMs.
% Then, we calculate the weight $W_i$ by interpolating the relative amplitude of pixel $(x,y)$ using a transfer function $T_i(x,y)$ that maps the relative resolution to relative amplitude. Let $i^*$ be the such that $R_{i^*}<R(x,y)<R_{i^*-1}$, then for all $i\in[1,4]$ we set
% $$ W_i(x,y) = 
% \begin{cases} 
%   B_i(x,y) & i=i^*-1 \\
%   1-B_i(x,y) & i=i^* \\
%   0 & \text{otherwise}
% \end{cases},$$
% where
% $$ B_i(x,y) = \frac{0.5-T_i(x,y)}{T_{i-1}(x,y)-T_i(x,y)}.$$
% The transfer function 
% % $T_i(r)=\exp\big(-(2^{i-2}r/\sigma)^2\big)$
% is defined as 
% $$T_i(x,y) = 
% \begin{cases} 
%     \exp\big(-\frac{(2^{i-2}R_i(x,y))^2}{2\sigma^2}\big) & i=1,\cdots,3\\
%     0 & i=4
% \end{cases},
% $$
In \cite{perry2002gaze}, $\alpha$ is tuned to match human perception via physiological experiments. Here we learn the parameters of FFMs, $\alpha$ and $\sigma$, together with the policy from human gaze data directly.

\subsection{Reward and Policy  Learning}\label{sec:policy}
Using FFMs as our state representation, we train a policy that mimics human gaze behavior using the IRL framework \cite{yang2020predicting}. However, we found that the GAIL \cite{ho2016generative}  IRL algorithm used in \cite{yang2020predicting} is too sensitive to its hyper-parameters, due to its adversarial learning design, which is also shown in \cite{kostrikov2018discriminatoractorcritic}. We therefore use IQ-Learn \cite{garg2021iq} as our IRL algorithm instead. Based on soft Q-Learning \cite{haarnoja2017reinforcement}, IQ-Learn encodes both the reward and the policy in a single Q-function, and thus is able to optimize both reward and policy simultaneously.%, in contrast to alternating between optimizing reward and policy in GAIL.

Let $Q(s,a)$ be the Q-function, which maps a state-action pair $(s,a)$ to a scalar value representing the amount of future reward gained by taking action $a$ under state $s$. We want to find a reward function that maximizes the expected amount of cumulative rewards that the expert policy obtains over all other possible policies. Hence, IQ-Learn trains the Q-function by minimizing the following loss:
\begin{align}
\mathcal{L}_\text{irl}=-\mathbb{E}_{\rho_E} \big[Q(s,a) -\gamma \mathbb{E}_{s'\sim\mathcal{P}(s,a)}V(s')\big],
\end{align}	
where $V(s)=\log\sum_a \exp(Q(s,a))$, $\rho_E$ and $\mathcal{P}$ denote the occupancy measure of the expert policy \cite{ho2016generative} and the dynamics, respectively. We do not apply the $\chi^2$-divergence proposed in \cite{garg2021iq} on the reward function since it did not lead to any notable   improvement on our task. Given the learned Q-function $Q$, we can compute the reward as a function of the state and action:
\begin{align}\label{eq:policy}
r(s,a)=Q(s,a)-\gamma\mathbb{E}_{s'\sim\mathcal{P}(s,a)}V(s'), 
\end{align} 
and the policy as a function of the state:
\begin{align}\pi(a|s)=\frac{\exp(Q(s,a)/\tau)}{\sum_{a'} \exp(Q(s,a')/\tau)}. \end{align}
$\tau$ is the temperature coefficient, controlling the entropy of the action distribution.

\myheading{Action space.} Our task is to predict the next fixation given the previous fixations, the input image, and the categorical target. To predict fixations on an image, we follow \cite{yang2020predicting} and discretize the image space into a $20{\times}32$ grid (action space). At each time step, the policy samples one cell out of 640 grid cells according to the predicted categorical action distribution $\pi(\cdot|s)$. For the selected grid cell, we set the predicted fixation to be the center of the cell.

\myheading{Auxiliary detection task. } A visual search task is essentially a detection task, so it is important for the state representation to capture features of the target object. Moreover, in target-absent search where the target object is absent, human behavior is driven by the expected location of the target in relation to other commonly co-occurring objects. In contrast to \cite{yang2020predicting} which directly uses the output of a pretrained panoptic segmentation network, we train the Q-function with an auxiliary task of predicting the center maps of the objects. Specifically, we add a detection network module on top of FFMs. This module outputs 80 heatmaps $\hat{Y}$ for the 80 object categories in the COCO dataset  \cite{lin2014microsoft}. Let $\hat{Y}_{xyc}$ denote the value of the $c$-th heatmap at location $(x,y)$. Following CenterNet~\cite{zhou2019objects}, we use pixel-wise focal loss \cite{lin2017focal} as an additional loss to train the whole network:
\begin{align}
\mathcal{L}_\text{det} = -\frac{1}{N}\sum_{x,y,c}
\begin{cases} 
    (1-\hat{Y}_{xyc})^\kappa\log(\hat{Y}_{xyc}) & \textrm{if } Y_{xyc}=1,\\
    (1-{Y}_{xyc})^\lambda(\hat{Y}_{xyc})^\kappa\log(1-\hat{Y}_{xyc}) & \text{otherwise,}
\end{cases}
\end{align}
where $Y$ is the ground-truth heatmap created by an object size dependent Gaussian kernel \cite{law2018cornernet}. We set $\kappa=2$ and $\lambda=4$ as in \cite{zhou2019objects}.
Note that we do not predict the exact heights and widths of the objects in the image because we think rough estimates of the locations of different objects are sufficient to help predict the target-absent fixations. We learn the Q-function using both the IRL loss and the auxiliary detection loss:
\begin{align}\label{eq:loss}
\mathcal{L}=\mathcal{L}_\text{irl} + \omega \mathcal{L}_\text{det},
\end{align}
where $\omega$ is a weight to balance the two loss terms.

\myheading{Termination Prediction.}
\label{sec:stop}
When  a person will stop searching is a question intrinsic to target-absent search. Different from \cite{chen2021predicting}, which formulates termination as an extra action to fixation prediction in policy learning, we treat termination prediction as an additional task that occurs every step after a new fixation has been made. We found that if we treat termination as an extra action, the policy would overfit to the termination action as it appears much more frequent than other actions.

To this end, we train a binary classifier on top of the Q-function (see \Sref{sec:policy}) for termination prediction using binary cross entropy loss. We weigh the loss computed on the termination and non-termination actions inversely proportionally to their frequencies. In addition, psychology studies \cite{chun1996just,wolfe2021guided} have suggested that time could be an important ingredient in predicting stopping. However, we do not predict the duration of fixations in our model. Instead, we use the number of previous fixations as an approximation of time and concatenate it with the Q-values from the Q-function as input to train the termination classifier.
% To test these hypothesis, we design the input to the termination predictor to be a combination of the following components: 1) maximum reward $r(s,a)$; 2) maximum Q value $Q(s,a)$; and 3) time. Since our model do not predict fixation durations, we use the number of fixations that has been made to approximate time. We compare the effect of each component and their combinations in \Sref{sec:ablation}.

\section{Experiments}
We train and evaluate the proposed method and other models by using COCO-Search18 \cite{chen2021coco}, which contains both target-present and target-absent human scanpaths in searching for 18 different object categories. COCO-Search18 has 3101 target-present images and 3101 target-absent images, each viewed by 10 subjects. In this paper, we mainly focus on the target-absent gaze behavior prediction. All models are only trained with target-absent images and fixations unless otherwise specified. For all models, we predict one scanpath for each testing image in a greedy manner (i.e., always selecting the action with the largest probability mass from the predicted action distribution as the next fixation) and compare them with the ground-truth scanpaths.

\subsection{Semantic Sequence Score}
The sequence score (SS) has often been used to quantify the success of scanpath prediction \cite{borji2013analysis,yang2020predicting}. The sequence score is computed by an existing string matching algorithm that compares the two fixation sequences \cite{needleman1970general} after transforming them into strings of fixation cluster IDs. The fixation clusters are computed based on the fixation locations. 
However, we argue that the sequence score does not capture the semantic meaning of fixations which plays an important role in analyzing goal-directed attention: it only captures ``where'' a person is looking at but not ``what'' is being looked at. To this end, we propose the {\it Semantic Sequence Score (SemSS)}, which transforms a fixation sequence into an {\it object category} sequence by leveraging the segmentation annotation provided in COCO \cite{lin2014microsoft}. Then, we apply the same string matching algorithm used in the traditional sequence score to measure the similarity between two scanpaths. Using the \enquote{things} versus \enquote{stuff} paradigm \cite{caesar2018coco}, we do not distinguish between object instances. In this paper, we focus on \enquote{thing} categories only, as we are interested in how non-target objects collectively affect human gaze behavior in visual search tasks. \enquote{Stuff} categories can be easily integrated into the semantic sequence score.

\myheading{Other metrics. }
We also report other scanpath prediction metrics including the traditional sequence score and conditional priority maps \cite{kummerer2021state}, which measure how well the model predicts a fixation when given the previous fixations using saliency metrics including information gain (IG) and normalized scanpath saliency (NSS) \cite{bylinskii2018different}. For clarity, we denote them by cIG and cNSS where \enquote{c} represents \enquote{conditional}. cIG measures the amount of information gain the model prediction has over a task-specific fixation density map computed using the training fixations. cNSS measures the correspondence between the predicted fixation probability map and the ground-truth fixation.
In addition, to measure  termination prediction accuracy, we report the Mean Absolute Error (MAE) between  predicted and ground-truth scanpath lengths.
% and the AUROC of (binary) termination classifier fed with the ground-truth fixations. 
To compare fairly with models that do not terminate automatically such as IRL \cite{yang2020predicting}, we also report the truncated sequence score by truncating predicted and ground-truth scanpaths at the first 2 and 4 new fixations, denoted as SS(2) and SS(4), respectively.

\subsection{Implementation Details}
\myheading{Network structure.}
Following \cite{yang2020predicting}, we resize the input images to $320\times 512$ for computational efficiency. As shown in \Fref{fig:ffm}, our model has three components: a set of $1\times1$ convolutional layers that project the feature maps in the feature pyramid to the same dimension (i.e., the number of channels in FFMs); an object detection module; and a fixation prediction module. We set the number of FFMs channels to 128.
The fixation prediction module and the object detection module share the same ConvNet consisting of three consecutive convolutional blocks which reduce the spatial resolution of the input foveated feature maps (FFMs) by a factor of 8 (from $160\times 256$ to $20\times 32$). In between two consecutive convolutional layers of a convolutional block, we apply Layer Normalization \cite{ba2016layer} and a ReLU activation function. Finally, the fixation prediction module uses two convolutional layers to map the output of the shared ConvNet into 18 attention maps (one for each target in COCO-Search18 \cite{chen2021coco}). The object detection module has a similar structure, but outputs 80 center maps (one for each object category in COCO \cite{lin2014microsoft}). Note that the backbone networks of all models in this paper are kept fixed during training. Detailed network parameters are in  supplementary.

\myheading{Hyperparameters.}
We train the models in this paper by using the Adam \cite{kingma2015adam} optimizer with learning rate $10^{-4}$. The weight for the auxiliary detection loss $\omega$ in \Eref{eq:loss} is 0.1.
In COCO-Search18 \cite{chen2021coco}, the number of pixels in one degree of visual angle $p=9.14$. We scale it according to the spatial resolution of $P_1$ and set $p=4.57$. For models with a termination predictor, we set the maximum length of each predicted scanpath to 10 (excluding the initial fixation) during training and testing. 
% This is longer than 90\% of the target-absent scanpaths in COCO-Search18. 
For models that do not terminate automatically, we set the length of the scanpath to 6 which is approximately the average length of the target-absent scanpaths in COCO-Search18.
For the IQ-Learn algorithm, the reward discount factor is set to 0.8. Following \cite{haarnoja2018soft,garg2021iq}, we use target updates and a replay buffer in IQ-Learn to stabilize the training. The temperature coefficient $\tau$ in \Eref{eq:policy} is set to 0.01. We update the target Q networks for four iterations using exponential moving average with a  0.01 coefficient. The replay buffer can hold 8000 state-action pairs and is updated online during training.

\subsection{Comparing Scanpath Prediction Methods}\label{sec:main_result}

We compare our model with the following baselines: 1) {\it human consistency}, an oracle method where one searcher's scanpath is used to predict another searcher's scanpath; 2) {\it detector}, a ConvNet trained on target-present images of COCO-Search18 to output a target detection confidence map, from which we sample fixations sequentially with inhibition of return (IOR); 3) {\it fixation heuristic}, similar to detector, but trained to predict human fixation density maps using target-absent data; and more recent approaches including 4) {\it IRL} \cite{yang2020predicting}, and 5) {\it Chen \etal's model} \cite{chen2021predicting}. Note that Chen \etal's model used a finer action space $30\times 40$. For fair comparison, we rescale its predicted fixations to our action space $20\times 32$.

\setlength{\tabcolsep}{6pt}
\begin{table}[t]
\begin{center}
\caption{{\bf Comparing target-absent scanpath prediction algorithms} (rows) using multiple scanpath metrics (columns) on the target-absent test set of COCO-Search18. The best results are highlighted in bold.}
\label{table:all_results}
\begin{tabular}{lcccccc}
\toprule
 &  SemSS  & SS & cIG  & cNSS  & SS(2) & SS(4)\\ \hline
Human consistency & 0.542 & 0.381 & - & - & 0.561 & 0.478 \\\hline
Detector & 0.497 & 0.321 & -0.516 & 0.446 & 0.497 & 0.402\\
Fixation heuristic & 0.484 & 0.298 & -0.599 & 0.405 & 0.492 & 0.379\\
IRL \cite{yang2020predicting} & 0.476 & 0.319 & 0.032 & 1.202 & 0.508 & 0.407\\
% CFI \cite{zelinsky2019benchmarking} \\
Chen \etal~\cite{chen2021predicting} & 0.484 & 0.331 & - & - & 0.516 & 0.434\\
Ours & {\bf 0.516} & {\bf 0.372} & {\bf 0.729} & {\bf 1.524} & {\bf 0.537}	& {\bf 0.441}\\
\bottomrule
\end{tabular}

\end{center}
\end{table}

As can be seen from \Tref{table:all_results}, our method outperforms all other methods across all metrics in target-absent scanpath prediction\footnote{Both cIG and cNSS can only be computed for auto-regressive probabilistic models (our method, IRL, detector and fixation heuristic).}. Our method is the closest to human consistency which is regarded as the ceiling of any predictive model. In the  sequence score case, our method is only inferior to human consistency by 0.09, leading the second best (Chen \etal~\cite{chen2021predicting}) by 0.41. Excluding the effect of the termination predictor, the sequence scores of the first 2 and 4 fixations also show that even without terminating the scanpaths our method is still the best compared to all other computational models. Moreover, comparing the sequence scores of truncated scanpaths and full scanpaths, we see a trend of decreasing performance as the scanpath length increases for all methods, i.e., SS(2) $>$ SS(4) $>$ SS, and this pattern is particularly pronounced in target-absent search (there is no significant difference between SS and SS(4) for target-present search, see \Tref{table:tp_rst}).
% This suggest that later fixations in a scanpath are harder to predict, which might indicate that {\it the target-absent gaze behavior is more random as the time spent on the search increases.}
{The fact that later fixations during target-absent search are harder to predict suggests that human eye-movements behave more randomly at the later stage of search especially when there is no target in the scene.}

\begin{figure}[t]
  \centering
  \includegraphics[width=1.0\linewidth]{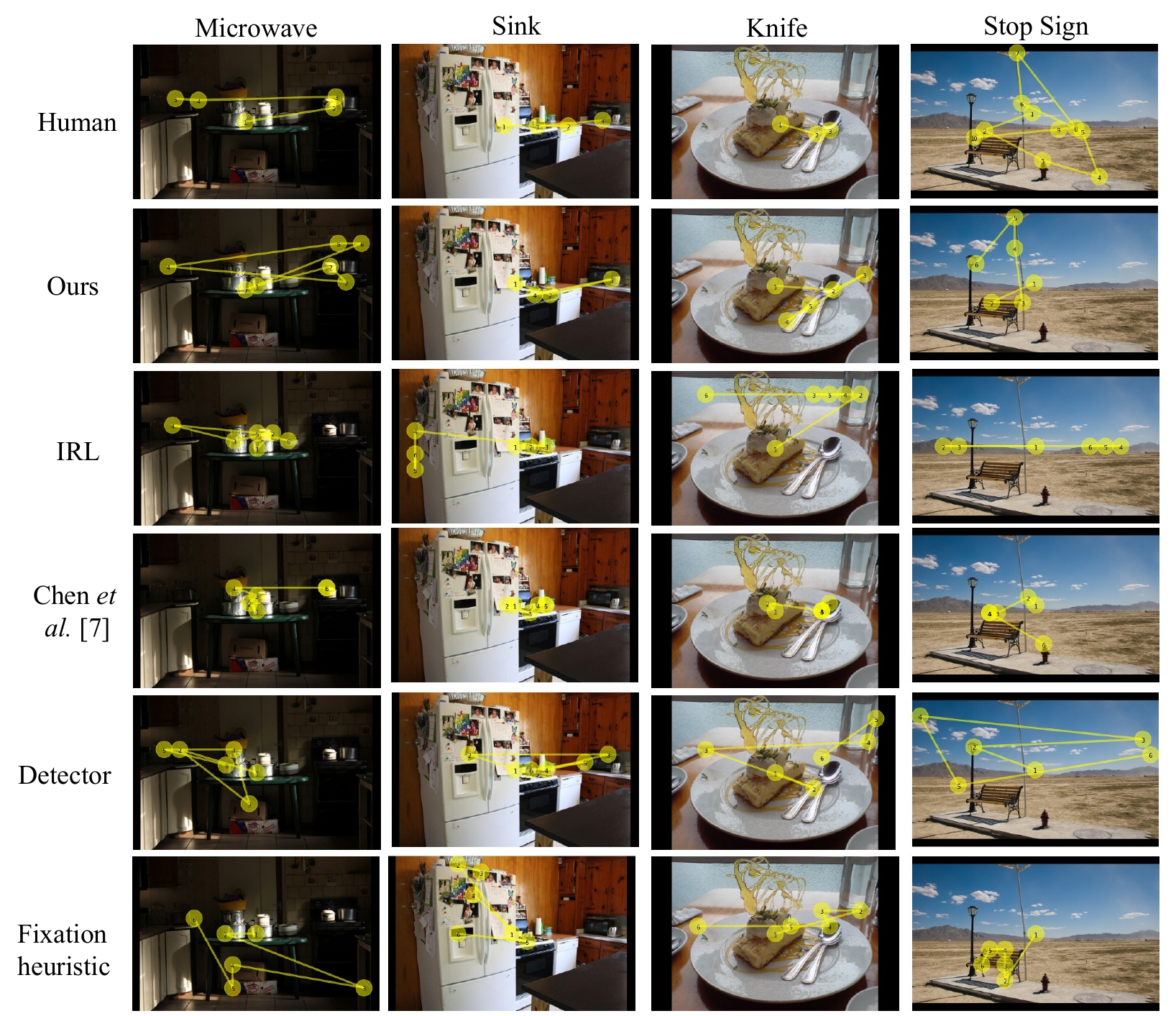}
%   \vskip -0.1in
  \caption{Visualizing the {\bf predicted scanpaths} of different methods (rows) for different search targets (columns). 
%   From left to right, the targets are microwave, sink, knife and stop sign. 
  The top row shows the ground-truth human scanpaths and the other rows are predicted scanpaths from different models.}
  \label{fig:vis}
\end{figure}

We also qualitatively compare different methods by visualizing their predicted scanpaths for four scenes in \Fref{fig:vis}. When searching for a microwave in this scene, our method alone predicted fixations on all  three table and countertop surfaces in the image where microwaves are often found (similar to how a representative human searched). Similar phenomena are observed for the sink and knife searches. This shows that our method is able to capture the contextual relations between objects that play a role in driving target-absent fixations. When searching for the stop sign, our method was the only one that looked at the top of the centrally-located vertical object, despite heavy occlusion, speculatively because stop signs are usually mounted to the tops of poles. In contrast, IRL, which extracts features from blurred pixels using a pretrained ConvNet, completely failed to capture the vertical objects in this image that seem to be guiding search. This argues for the value in using our proposed FFMs to capture guiding contextual information extracted from peripheral vision.

\subsection{Group Model versus Individual Model}
In target-present search, human scanpaths are very consistent due to the strong guidance provided by the target object in the image. Indeed, 
% previous study on target-present search prediction showed that the 
a model trained with fixations from a group of people generalized well for a new {\bf unseen} person \cite{yang2020predicting}. However, given that there are large individual differences in termination time for target-absent search~\cite{chen2021coco}, we expect that individualized modeling may be necessary for target-absent search prediction. To test this hypothesis, we compared the predictive performance of group versus individual modeling of target-absent search fixations. The group model was trained with 9 subjects' training scanpaths and tested on the testing scanpaths of the remaining subject. The individual model was trained with the training scanpaths of a single subject and tested on the same subject's testing scanpaths. We did this for all 10 subjects.% in the dataset.% and averaged the performance measures.

\begin{figure}[t]
  \centering
  \includegraphics[width=.95\linewidth]{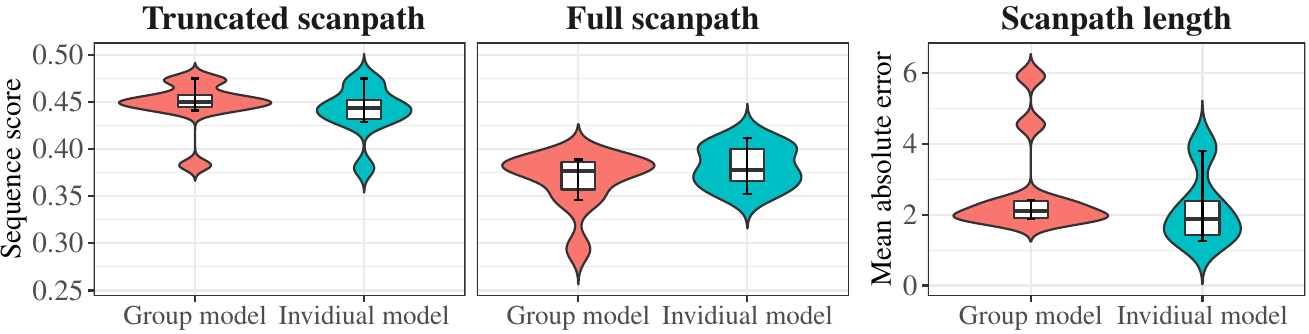}
  \caption{{\bf Comparing group model (red) and individual model (cyan)} using:  (left) the sequence score of the truncated scanpath (first 4 fixations) without automatic termination, and (middle) the sequence score of the full scanpath including termination, and (right) the mean absolute error for the predicted scanpath length.
  We perform Wilcoxon signed-rank tests for each experimental setting. The two-sided p values are 0.012, 0.028 and 0.006, respectively.
  }
  \label{fig:group_vs_indi}
\end{figure}

\Fref{fig:group_vs_indi} shows the comparison between the group model and the individual model in the sequence score of full scanpaths and truncated scanpaths (first four fixations) and the MAE of the length of the predicted scanpath.
% Interestingly, the performance of group model and the individual model are very similar in their first 4 fixations, but the individual model is significantly better in full-scanpath sequence score, which includes the effect of automatic scanpath termination. This suggests that the {\it human gaze behavior generalizes from one subject to another only in the fixation allocation but not in their termination criterion} evidenced by the fact that the individual model has a smaller MAE in terminating the scanapths in \Fref{fig:group_vs_indi}. 
Interestingly, despite being trained with less data, the individual model shows better performance than the group model in full scanpath modeling, contrary to the group model being better in the truncated scanpath prediction. A critical difference between the modeling of truncated versus full scanpath is that the latter involves the search termination prediction. The rightmost graph in \Fref{fig:group_vs_indi} also shows that the individual model generates less error (in MAE metric) in scanpath length prediction than the group model. These results altogether suggest that individualized modeling may be more suitable for target-absent search prediction. More experimental results on the termination criterion across different subjects can be found in supplementary.

\subsection{Ablation Study}\label{sec:ablation}

\setlength{\tabcolsep}{3pt}
\begin{table}[t]
\begin{center}
\caption{ {\bf Ablation study}. We ablate the loss function (second row) and the state representation (third and fourth rows). All methods are trained using IQ-Learn. \label{table:ablate_rst}
}
% \vskip -0.1in
\begin{tabular}{lcccccc}
\toprule
& SemSS & SS &  cIG  & cNSS  & SS(2) & SS(4)\\ \hline
FFMs & {\bf 0.516} & {\bf 0.372} & {\bf 0.729} & {\bf 1.524} & 0.537 & {\bf 0.441}\\
FFMs w/o detection loss & 0.476 & 0.350 & 0.550 & 1.332 & {\bf 0.545}	& 0.437\\
% \midrule
DCBs & 0.508 & 0.355 & 0.212 & 1.129 & 0.514	& 0.426 \\
CFI & 0.504 & 0.352 & 0.518 & 1.252 & 0.506 & 0.426\\
FPN & 0.508 &0.338 &0.018 &0.881 &0.408& 0.351\\
Binary masks & 0.510 & 0.364 & 0.347 & 1.148&  0.438 & 0.378\\
\bottomrule
\end{tabular}

\end{center}
\end{table}

First, we ablate the loss (see \Eref{eq:loss}) of our model by removing the auxiliary detection loss. Second, we ablate our proposed foveated feature maps (FFMs) by comparing it with dynamic contextual beliefs (DCBs) \cite{yang2020predicting} and cumulative foveated image (CFI) \cite{zelinsky2019benchmarking} using the same IRL algorithm (i.e., IQ-Learn).
As a finer-grained ablation, we ablate FFMs by using the features extracted by the FPN backbone of a COCO-pretrained Mask R-CNN as the state representation. We use the highest-resolution feature maps of FPN $P_2$. We further binarize the FFMs of our model such that the values are one at the fixated locations of the finest level (fovea) and the non-fixated locations of the coarsest level (periphery) and zero elsewhere.
As shown in \Tref{table:ablate_rst}, the proposed auxiliary detection loss improves the performance in 5 out of 6 metrics. The semantic sequence score is increased from 0.476 to 0.516, which indicates that knowing the locations of non-target objects in the image is helpful for predicting the target-absent fixations. 
Comparing different state representations (i.e., FFMs, DCBs, CFI, FPN and binary masks), we can see that the proposed FFMs are superior to all other state presentations in predicting target-absent fixations. This shows the superiority of FFMs in representing the knowledge a human acquires through fixations compared to DCBs and CFI, which apply pretrained ConvNets on blurred images to simulate the foveated retina.

% {\bf Termination prediction. }
% Lastly, we ablate the input of the termination classifier to see the effect of each components of the input to termination prediction \Sref{sec:stop}.

% \begin{table}[t]
% \begin{center}

% \begin{tabular}{l|ccc|cc}
% \toprule
%  & Q-value & Reward & Time & MAE & AUROC \\ \hline
% (a) & \checkmark & - & -\\
% (b) & - & \checkmark & -\\
% (c) & - & - & \checkmark\\
% (d) & \checkmark & - & \checkmark\\
% (d) & - & \checkmark & \checkmark\\
% \bottomrule
% \end{tabular}

% \end{center}
% \caption{ {\bf Ablating the termination classifier}. }
% \label{table:ablate_rst}
% \end{table}

\subsection{Generalization to Target-present Search}

\setlength{\tabcolsep}{6pt}
\begin{table}[t]
\begin{center}
\caption{ {\bf Comparing target-present scanpath prediction algorithms}  using multiple scanpath metrics on the COCO-Search18 test dataset.}
\label{table:tp_rst}
\begin{tabular}{lcccccc}
\toprule
 &  SemSS  & SS & cIG  & cNSS  & SS(2) & SS(4)\\ \hline
Human consistency & 0.624 & 0.478 & - & - & 0.486 & 0.480\\\hline
IRL \cite{yang2020predicting} & 0.536 & 0.419 & -9.709	& 1.977 & 0.437 & 0.421 \\
Chen \etal~\cite{chen2021predicting} & {\bf0.572} & 0.445 & - & - & {\bf 0.516} & {0.429}	\\
Ours & 0.562 & {\bf 0.451} & {\bf1.548} & {\bf2.376} & {0.467} & {\bf 0.450}\\
\bottomrule
\end{tabular}

\end{center}
\end{table}

Despite being motivated by target-absent search, our method is also directly applicable to target-present fixation prediction. In this section, we compare our model with two competitive models, IRL \cite{yang2020predicting} and Chen \etal~\cite{chen2021predicting}, in target-present scanpath prediction. For fair comparison, we follow \cite{yang2020predicting} and set the maximum scanpath length to be 6 (excluding the first fixation) for all models and automatically terminate the scanpath once the fixation falls in the bounding box of the target.
\Tref{table:tp_rst} shows that our method achieves the best performance in 4 out of 6 metrics. Chen \etal's model is slightly better than ours in semantic sequence score. They used a pretrained CenterNet~\cite{zhou2019objects} trained on COCO images \cite{lin2014microsoft} (about 118K images) to predict the bounding box of the target as input for their model, whereas we only used the target-present images in COCO-Search18 \cite{chen2021coco} (about 3K images) to train our object detection module (see \Fref{fig:ffm}). Despite being trained with less data, our model still outperforms Chen \etal~\cite{chen2021predicting} in the other five metrics, especially when evaluated in truncated fixed-length scanpaths (i.e., SS(2) and SS(4)).  We further expect our model to perform better when using all COCO training images to train our object detection module. \Tref{table:all_results} and \Tref{table:tp_rst} together demonstrate that our proposed method not only excels in predicting target-absent fixations (see \Sref{sec:main_result}), but also target-present fixations.

\section{Conclusions and Discussion}
We have presented the first computational model for predicting target-absent search scanpaths. To represent the internal knowledge that the viewer acquires through fixations, we proposed a novel state representation, {\it foveated feature maps (FFMs)}. FFMs circumvent the drawbacks of directly applying pretrained ConvNets on blurred images in previous methods \cite{zelinsky2019benchmarking,yang2020predicting} by integrating the in-network feature pyramid produced by a pretrained ConvNet with a foveated retina. When trained and evaluated on the COCO-Search18 dataset, FFMs outperform previous state representations and achieve state-of-the-art performance in predicting both target-absent and target-present search fixations using the IRL framework. Moreover, we also proposed a new variant of the sequence score for measuring scanpath similarity, called semantic sequence score. It better captures the object-to-object relation used to guide target-absent search.

% \myheading{Limitations.}
% We assume a perfect memory in the current form of our state representation. Namely, all high-resolution information collected through previous fixations stays in the current state for predicting the next fixation. On the contrary, humans tend to forget information acquired from previous fixations in visual search \cite{horowitz1998visual}.
% Equipping our FFMs with fading memory might be helpful in target-absent scanpath prediction as in target-absent search, subjects are more likely to return to a fixated location as memory about the fixated locations fades.

\myheading{Future work.}
Inspired by \cite{yang2020predicting}, our future work will involve extending our model and semantic sequence score to include \enquote{stuff} categories in COCO \cite{caesar2018coco} to study the impact of background categories to target-absent search gaze behavior, and exploring using semi-supervised learning to address the lack of  human gaze data by leveraging the rich annotation in COCO images \cite{lin2014microsoft}.

\myheading{Acknowledgements.}
The authors would like to thank Jianyuan Deng for her help in result visualization and statistical analysis. This project was partially supported by US National Science Foundation Awards IIS-1763981 and IIS-2123920, the Partner University Fund, the SUNY2020 Infrastructure Transportation Security Center, and a gift from Adobe.
\clearpage
\bibliographystyle{splncs04}
\bibliography{egbib}
\end{document}

%% file: definitions.tex
\def\mA{\mathcal{A}}
\def\mB{\mathcal{B}}
\def\mC{\mathcal{C}}
\def\mD{\mathcal{D}}
\def\mE{\mathcal{E}}
\def\mF{\mathcal{F}}
\def\mG{\mathcal{G}}
\def\mH{\mathcal{H}}
\def\mI{\mathcal{I}}
\def\mJ{\mathcal{J}}
\def\mK{\mathcal{K}}
\def\mL{\mathcal{L}}
\def\mM{\mathcal{M}}
\def\mN{\mathcal{N}}
\def\mO{\mathcal{O}}
\def\mP{\mathcal{P}}
\def\mQ{\mathcal{Q}}
\def\mR{\mathcal{R}}
\def\mS{\mathcal{S}}
\def\mT{\mathcal{T}}
\def\mU{\mathcal{U}}
\def\mV{\mathcal{V}}
\def\mW{\mathcal{W}}
\def\mX{\mathcal{X}}
\def\mY{\mathcal{Y}}
\def\mZ{\mathcal{Z}}

\def\1n{\mathbf{1}_n}
\def\0{\mathbf{0}}
\def\1{\mathbf{1}}

\def\A{{\bf A}}
\def\B{{\bf B}}
\def\C{{\bf C}}
\def\D{{\bf D}}
\def\E{{\bf E}}
\def\F{{\bf F}}
\def\G{{\bf G}}
\def\H{{\bf H}}
\def\I{{\bf I}}
\def\J{{\bf J}}
\def\K{{\bf K}}
\def\L{{\bf L}}
\def\M{{\bf M}}
\def\N{{\bf N}}
\def\O{{\bf O}}
\def\P{{\bf P}}
\def\Q{{\bf Q}}
\def\R{{\bf R}}
\def\S{{\bf S}}
\def\T{{\bf T}}
\def\U{{\bf U}}
\def\V{{\bf V}}
\def\W{{\bf W}}
\def\X{{\bf X}}
\def\Y{{\bf Y}}
\def\Z{{\bf Z}}

\def\a{{\bf a}}
\def\b{{\bf b}}
\def\c{{\bf c}}
\def\d{{\bf d}}
\def\e{{\bf e}}
\def\f{{\bf f}}
\def\g{{\bf g}}
\def\h{{\bf h}}
\def\i{{\bf i}}
\def\j{{\bf j}}
\def\k{{\bf k}}
\def\l{{\bf l}}
\def\m{{\bf m}}
\def\n{{\bf n}}
\def\o{{\bf o}}
\def\p{{\bf p}}
\def\q{{\bf q}}
\def\r{{\bf r}}
\def\s{{\bf s}}
\def\t{{\bf t}}
\def\u{{\bf u}}
\def\v{{\bf v}}
\def\w{{\bf w}}
\def\x{{\bf x}}
\def\y{{\bf y}}
\def\z{{\bf z}}

\def\balpha{\mbox{\boldmath{$\alpha$}}}
\def\bbeta{\mbox{\boldmath{$\beta$}}}
\def\bdelta{\mbox{\boldmath{$\delta$}}}
\def\bgamma{\mbox{\boldmath{$\gamma$}}}
\def\blambda{\mbox{\boldmath{$\lambda$}}}
\def\bsigma{\mbox{\boldmath{$\sigma$}}}
\def\btheta{\mbox{\boldmath{$\theta$}}}
\def\bomega{\mbox{\boldmath{$\omega$}}}
\def\bxi{\mbox{\boldmath{$\xi$}}}
\def\bnu{\mbox{\boldmath{$\nu$}}}                                  
\def\bphi{\mbox{\boldmath{$\phi$}}}
\def\bmu{\mbox{\boldmath{$\mu$}}}

\def\bDelta{\mbox{\boldmath{$\Delta$}}}
\def\bOmega{\mbox{\boldmath{$\Omega$}}}
\def\bPhi{\mbox{\boldmath{$\Phi$}}}
\def\bLambda{\mbox{\boldmath{$\Lambda$}}}
\def\bSigma{\mbox{\boldmath{$\Sigma$}}}
\def\bGamma{\mbox{\boldmath{$\Gamma$}}}

\newcommand{\myminimum}[1]{\mathop{\textrm{minimum}}_{#1}}
\newcommand{\mymaximum}[1]{\mathop{\textrm{maximum}}_{#1}}    
\newcommand{\mymin}[1]{\mathop{\textrm{minimize}}_{#1}}
\newcommand{\mymax}[1]{\mathop{\textrm{maximize}}_{#1}}
\newcommand{\mymins}[1]{\mathop{\textrm{min.}}_{#1}}
\newcommand{\mymaxs}[1]{\mathop{\textrm{max.}}_{#1}}  
\newcommand{\myargmin}[1]{\mathop{\textrm{argmin}}_{#1}} 
\newcommand{\myargmax}[1]{\mathop{\textrm{argmax}}_{#1}} 
\newcommand{\myst}{\textrm{s.t. }}

\newcommand{\denselist}{\itemsep -1pt}
\newcommand{\sparselist}{\itemsep 1pt}

\definecolor{pink}{rgb}{0.9,0.5,0.5}
\definecolor{purple}{rgb}{0.5, 0.4, 0.8}   
\definecolor{gray}{rgb}{0.3, 0.3, 0.3}
\definecolor{mygreen}{rgb}{0.2, 0.6, 0.2}

\newcommand{\cyan}[1]{\textcolor{cyan}{#1}}
\newcommand{\red}[1]{\textcolor{red}{#1}}  
\newcommand{\blue}[1]{\textcolor{blue}{#1}}
\newcommand{\magenta}[1]{\textcolor{magenta}{#1}}
\newcommand{\pink}[1]{\textcolor{pink}{#1}}
\newcommand{\green}[1]{\textcolor{green}{#1}} 
\newcommand{\gray}[1]{\textcolor{gray}{#1}}    
\newcommand{\mygreen}[1]{\textcolor{mygreen}{#1}}    
\newcommand{\purple}[1]{\textcolor{purple}{#1}}       

\definecolor{greena}{rgb}{0.4, 0.5, 0.1}
\newcommand{\greena}[1]{\textcolor{greena}{#1}}

\definecolor{bluea}{rgb}{0, 0.4, 0.6}
\newcommand{\bluea}[1]{\textcolor{bluea}{#1}}
\definecolor{reda}{rgb}{0.6, 0.2, 0.1}
\newcommand{\reda}[1]{\textcolor{reda}{#1}}

\def\changemargin#1#2{\list{}{\rightmargin#2\leftmargin#1}\item[]}
\let\endchangemargin=\endlist
                                               
\newcommand{\cm}[1]{}

\newcommand{\mtodo}[1]{{\color{red}$\blacksquare$\textbf{[TODO: #1]}}}
\newcommand{\myheading}[1]{\vspace{1ex}\noindent \textbf{#1}}
\newcommand{\htimesw}[2]{\mbox{$#1$$\times$$#2$}}

\newcommand{\young}[1]{{\color{blue}$\blacksquare$\textbf{Alternative}: #1}}

% The following are useful for creating homework or exams

\newif\ifshowsolution
%\showsolutionfalse
\showsolutiontrue

\ifshowsolution  
\newcommand{\Comment}[1]{\paragraph{\bf $\bigstar $ COMMENT:} {\sf #1} \bigskip}
\newcommand{\Solution}[2]{\paragraph{\bf $\bigstar $ SOLUTION:} {\sf #2} }
\newcommand{\Mistake}[2]{\paragraph{\bf $\blacksquare$ COMMON MISTAKE #1:} {\sf #2} \bigskip}
\else
\newcommand{\Solution}[2]{\vspace{#1}}
\fi

\newcolumntype{L}[1]{>{\raggedright\let\newline\\\arraybackslash\hspace{0pt}}m{#1}}
\newcolumntype{C}[1]{>{\centering\let\newline\\\arraybackslash\hspace{0pt}}m{#1}}
\newcolumntype{R}[1]{>{\raggedleft\let\newline\\\arraybackslash\hspace{0pt}}m{#1}}

\newcommand{\truefalse}{
\begin{enumerate}
	\item True
	\item False
\end{enumerate}
}

\newcommand{\yesno}{
\begin{enumerate}
	\item Yes
	\item No
\end{enumerate}
}

\newcommand{\Sref}[1]{Sec.~\ref{#1}}
\newcommand{\Eref}[1]{Eq.~(\ref{#1})}
\newcommand{\Fref}[1]{Fig.~\ref{#1}}
\newcommand{\Tref}[1]{Tab.~\ref{#1}}
\newcommand{\mhoai}[1]{{\color{magenta}\textbf{[MH: #1]}}}